\documentclass[runningheads,orivec]{llncs}
\usepackage[T1]{fontenc}
\usepackage{amsmath}
\usepackage{graphicx}
\usepackage{booktabs}
\usepackage{csquotes}
\usepackage{xcolor}

\usepackage[colorlinks=true, allcolors=blue]{hyperref}
\usepackage[capitalize,nameinlink]{cleveref}

\usepackage[nolist]{acronym}
\begin{acronym}
    \acro{rl}[RL]{reinforcement learning}
    \acro{ppo}[PPO]{proximal policy optimization}
    \acro{drl}[DRL]{deep reinforcement learning}
    \acro{mdp}[MDP]{Markov decision process}
    \acro{mpc}[MPC]{model predictive control}
    \acro{1dof}[1-DoF]{one-degree-of-freedom}
    \acro{lqr}[LQR]{linear-quadratic regulator}
    \acro{lqi}[LQI]{öinear-quadratic-integral regulator}
    \acro{lti}[LTI]{linear Time-Invariant}
    \acro{sb3}[SB3]{Stable Baselines3}
    \acro{trpo}[TRPO]{trust region policy optimization}
    \acro{icps}[ICPS]{industrial cyber-physical system}
    \acro{cps}[CPS]{cyber-physical system}
    \acro{pid}[PID]{proportional-integral-derivative}
\end{acronym}

\begin{document}
\title{Anticipatory Reinforcement Learning for Trajectory Tracking}
\titlerunning{Anticipatory RL for Trajectory Tracking}
% If the paper title is too long for the running head, you can set
% an abbreviated paper title here
%
\author{Georg Schäfer\inst{1,2} \and Jakob Rehrl\inst{1} \and Stefan Huber\inst{1} \and Simon Hirlaender\inst{2}
}
\authorrunning{G. Schäfer et al.}
% First names are abbreviated in the running head.
% If there are more than two authors, 'et al.' is used.
%
\institute{Josef Ressel Centre for Intelligent and Secure Industrial Automation, \\
Salzburg University of Applied Sciences, Salzburg, Austria \and
Paris Lodron University of Salzburg, Salzburg, Austria \\
\email{georg.schaefer@fh-salzburg.ac.at}}
\maketitle              % typeset the header of the contribution
\begin{abstract}

Deep reinforcement learning (DRL) in industrial control often suffers from lag and overshoot due to purely reactive control based on the current tracking error.
To achieve anticipatory control without high computational overhead, we introduce a predictive formulation that augments the DRL state space with target velocities and future reference horizons.
Evaluating eight configurations using proximal policy optimization (PPO) on a 1-degree-of-freedom (1-DoF) helicopter testbed, simulation results showed a 9-fold error reduction, lowering the mean absolute deviation from $2.73^\circ$ to $0.31^\circ$.
However, zero-shot transfer to physical hardware revealed a sim-to-real gap.
Interestingly, a simpler configuration using a single, further look-ahead horizon matched the real-world top performance of the most complex model ($1.11^\circ$).
Overall, evaluating various combinations of prediction horizons and target velocities demonstrated that highly granular predictive data is not necessarily required for physical transfer.

\end{abstract}
\section{Introduction}

\Acp{icps} integrate physical processes with computational control and underpin modern manufacturing, energy management, and smart production lines.
\Ac{rl} has demonstrated success in optimizing these complex, dynamic environments, offering data-driven control solutions where mathematical modeling is challenging~\cite{nian2020review}.
However, many \ac{drl} policies typically compute their actions based on the current state and the instantaneous reference, without anticipating future changes in the reference trajectory.
For example, standard benchmark environments and baseline controllers often rely solely on the instantaneous deviation from a target (e.g., \cite{brockman2016openai,schafer2025crucial}).
Consequently, these agents frequently exhibit lag and overshoot, as they cannot anticipate upcoming trajectory changes. In contrast, classic control strategies such as \ac{mpc} systematically incorporate future reference trajectories. However, computing optimal actuating sequences requires solving an optimization problem at every time step. This high computational demand limits \ac{mpc} in real-time industrial applications.
\Ac{drl} offers computationally lightweight online inference once trained.
To bridge the gap between the foresight of \ac{mpc} and the computational efficiency of \ac{drl}, this work-in-progress paper introduces a predictive problem formulation for industrial mechatronic systems.
Our primary contributions are twofold:
(i) We embed \enquote{foresight} into the \ac{drl} state space by augmenting the agent's observation with future target variables and the current target velocity.
(ii) We systematically evaluate the sim-to-real transferability of these predictive agents on a physical \ac{1dof} helicopter testbed to assess their performance with regards to reference tracking.

\section{Problem Formulation and Experimental Setup}
The experimental setup utilizes the Quanser Aero 2 testbed in a \ac{1dof} configuration.
We access the simulation and the physical system as described in~\cite{schafer2024python}.
Building upon our previous problem formulation \cite{schafer2025crucial}, we redefine the tracking problem to better align with the dynamic constraints of physical hardware.
Instead of step functions as reference signals, which are physically impossible to track for mechanical systems, we shift the target generation to continuous S-curve trajectories and constant sections.
These S-curves provide physically feasible, differentiable trajectories.
This prevents the agent from overfitting its policy to the unrealistic, instantaneous error jumps caused by rigid step functions, forcing it instead to learn generalized, dynamic tracking behavior.
In a purely reactive setup, the environment state at time $t$ is defined as $s_t = (\theta_t, \omega_t, r_t)$, where $\theta_t$ is the pitch angle, $\omega_t$ is the angular velocity (computed as $\Delta\theta/\Delta t$), and $r_t$ is the current reference.
To embed predictive \enquote{foresight}, we augment this state space by including the target velocity $\dot{r}_t$ (computed analytically) and a horizon of future target points sampled directly from the trajectory.
The augmented state space is formulated as $s_t = (\theta_t, \omega_t, r_t, \dot{r}_t, r_{t+1\cdot\Delta t}, ..., r_{t+N\cdot\Delta t})$.
Here, $N$ represents the number of future targets provided to the agent, and $\Delta t$ defines the fixed time interval between these future trajectory points.
To solve this control task, we selected the \ac{ppo} algorithm due to its theoretical policy improvement guarantees and proven stability in continuous action spaces.
The algorithm was implemented using the Stable Baselines3 library~\cite{raffin2021stable}.
To isolate the impact of predictive features, we systematically evaluated eight state-space configurations formed by augmenting the purely reactive baseline with various combinations of the target velocity $\dot{r}$ and look-ahead horizons ($N$ future targets at varying intervals $\Delta t$).
(2) \textbf{Target Velocity:} augmented solely with $\dot{r}_t$; \textbf{Future Targets:} augmented with $N$ future targets at varying intervals of $\Delta t$;
and (4) \textbf{Combined:} integrating both $\dot{r}_t$ and future targets.

To ensure statistical significance, each configuration was trained in simulation over 10 independent runs using varying random seeds.
The primary evaluation metric was the mean absolute deviation to the target trajectory.
Finally, to evaluate the sim-to-real gap, the single best-performing agent from each configuration was deployed directly onto the physical hardware and evaluated on a single representative S-curve trajectory.

\section{Results and Discussion}
\begin{figure}[htbp]
    \centering
    \includegraphics[width=\textwidth]{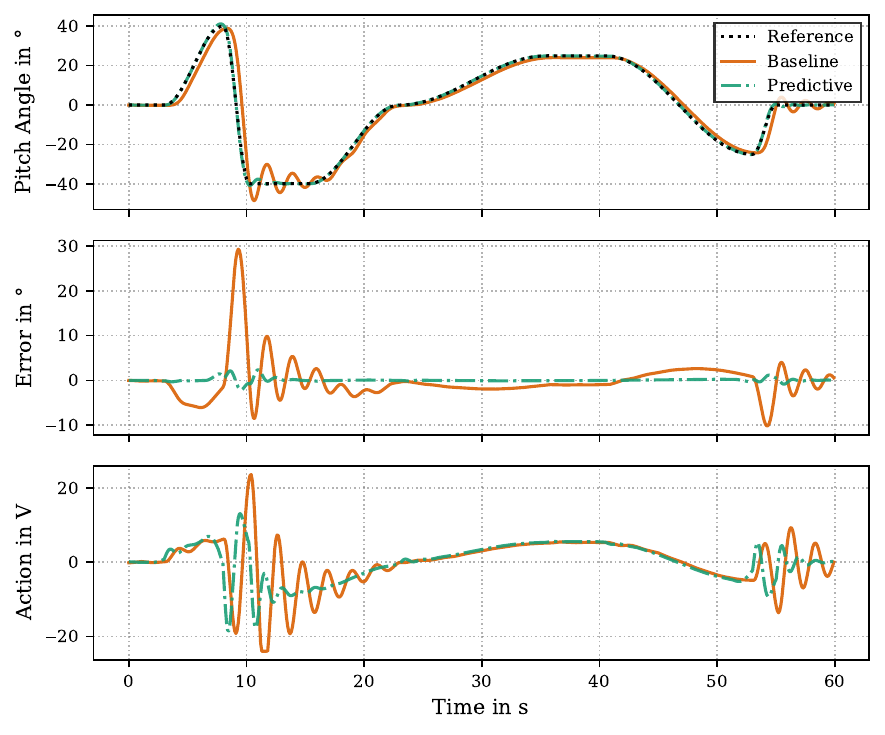}
    \caption{Simulation evaluation profile of the reactive baseline versus the best predictive agent, showing the S-curve pitch tracking (top), target deviation error (middle), and applied control voltage (bottom).}
    \label{fig:tracking_profile}
\end{figure}

The experimental results of all eight evaluated configurations are summarized in \cref{tab:predictive_results}.
In the simulation environment, the introduction of predictive state variables yielded a roughly 9-fold improvement in tracking accuracy.
The purely reactive baseline agent struggled to track the continuous S-curves, resulting in a mean absolute deviation of $2.73^\circ \pm 0.18^\circ$.
In contrast, the fully combined predictive agent (utilizing both the target velocity $\dot{r}_t$ and $N=4$ future targets at $\Delta t = 0.5$\,s intervals) achieved near-perfect tracking with a deviation of just $0.31^\circ \pm 0.10^\circ$.
A critical finding from the simulation data is the impact of the look-ahead horizon length.
When providing the agent with only a single future target ($N=1$), a time interval of $\Delta t = 1.0$\,s significantly outperformed a short $\Delta t = 0.1$\,s interval ($0.85^\circ \pm 0.06^\circ$ versus $1.70^\circ \pm 0.13^\circ$, respectively).
This behavior can be attributed to the physical inertia of the \ac{1dof} system.
A $0.1$\,s look-ahead does not provide sufficient time for the agent to accelerate or decelerate the beam.
Conversely, a $1.0$\,s horizon allows the agent to smoothly \enquote{pre-steer} and apply control efforts proactively, minimizing lag.
To visualize this improvement \cref{fig:tracking_profile} illustrates the tracking behavior of the baseline agent compared to the predictive agent against the reference signal.

While the simulation results demonstrate the significant improvement of the predictive formulation, the real-world validation on the physical hardware reveals a classic sim-to-real gap. The most complex agent ($\dot{r}_t$, $N=4$ with $\Delta t = 0.5$\,s) performed best in simulation.
However, during physical hardware validation, this most augmented configuration did not outperform simpler models; instead, it yielded the same top-tier performance as an agent with no target velocity and only a single future target with $\Delta t = 1.0\,s$.
This phenomenon provides valuable insights into mitigating the sim-to-real gap.
By providing the agent with highly granular, dense trajectory data, the neural network is able to exploit the idealized, noise-free dynamics of the simulation model.
However, physical systems suffer from unmodeled friction, communication delays, and sensor noise.
As observed in recent literature regarding the sim-to-real gap \cite{wagenmaker2024overcoming}, policies that perfectly optimize for simulated dynamics often become fragile in reality.
Because the simpler predictive formulation matches the hardware performance of the most complex agent, it becomes clear that highly dense trajectory data is not strictly required for optimal physical transfer.
Instead, a simpler, broader look-ahead horizon acts as an effective regularizer.
It provides enough foresight to overcome the system's inertia while remaining abstract enough to ignore minor dynamic discrepancies between the simulation and the real-world testbed, offering a more computationally efficient state space without sacrificing real-world accuracy.

\begin{table}[htbp]
    \centering
    \caption
    {Comparison of tracking performance across different predictive state representations. \textbf{$\dot{r}$}: Target velocity included. \textbf{$N$}: Number of future target points. \textbf{$\Delta t$}: Time interval between future targets in seconds. Values represent the absolute pitch deviation in degrees~($^\circ$). Worst, Best, Mean, and Std refer to the maximum, minimum, average, and standard deviation of the tracking error across the 10 independent simulation runs.}
    \label{tab:predictive_results}
    \setlength{\tabcolsep}{12pt}
    \begin{tabular}{@{}ccc|cccc|c@{}}
        \toprule
        \textbf{$\dot{r}$} & \textbf{$N$} & \textbf{$\Delta t$} & \textbf{Worst} & \textbf{Best} & \textbf{Mean} & \textbf{Std} & \textbf{Hardware} \\
        \midrule
        No  & 0 & --  & 3.05 & 2.51 & 2.73 & 0.18 & 2.34 \\
        No  & 1 & 0.1 & 1.88 & 1.49 & 1.70 & 0.13 & 2.09 \\
        No  & 1 & 1.0 & 0.95 & 0.78 & 0.85 & 0.06 & \textbf{1.11} \\
        No  & 4 & 0.5 & 1.02 & 0.49 & 0.70 & 0.14 & 1.21 \\
        \midrule
        Yes & 0 & --  & 0.90 & 0.41 & 0.52 & 0.15 & 1.23 \\
        Yes & 1 & 0.1 & 0.74 & 0.34 & 0.48 & 0.10 & 1.43 \\
        Yes & 1 & 1.0 & 0.50 & 0.22 & 0.34 & 0.08 & 1.14 \\
        Yes & 4 & 0.5 & 0.53 & 0.19 & \textbf{0.31} & 0.10 & \textbf{1.11} \\
        \bottomrule
    \end{tabular}
\end{table}

\section{Conclusion and Future Work}
We demonstrated that the performance limitations of standard \ac{drl} in physical tracking stem largely from their purely reactive problem formulation.
Embedding a predictive look-ahead horizon into the state space bridges the gap between reactive AI and optimal control.
This augmented formulation allows the agent to anticipate trajectory changes and compensate for physical inertia, resulting in highly accurate continuous tracking.
For industry, this approach offers a highly desirable trade-off: lightweight online inference of neural networks while maintaining the high precision required to stabilize complex plants.
Furthermore, sim-to-real validation highlighted that a coarser prediction horizon acts as a robust regularizer, preventing overfitting to simulated dynamics.
Future work will expand this approach.
First, to mitigate the reliance on \textit{a priori} target trajectories, we will explore learning trajectory prediction models for task-specific movement patterns via supervised learning.
Second, we plan fine-tuning directly on hardware and conducting statistically rigorous hardware evaluations across a broader $(N, \Delta t)$ grid.
Third, we aim to scale this predictive formulation to multi-DoF robotic systems for complex manufacturing tasks. Finally, we will benchmark the real-time performance and computational overhead of our agents against advanced \ac{mpc}.

\paragraph*{Acknowledgments.}{
Financial support for this study was provided by the Christian Doppler Research Association (CDG) through the Josef Ressel Centre for Intelligent and Secure Industrial Automation, the corresponding WISS Co-project of Land Salzburg, and by the European Interreg project BA0100172 AI4GREEN.
During the preparation of this work, the authors used AI tools for language editing and formatting assistance.
}

%
% ---- Bibliography ----
%
% BibTeX users should specify bibliography style 'splncs04'.
% References will then be sorted and formatted in the correct style.
%
\bibliographystyle{splncs04}
\bibliography{references}

\end{document}